\documentclass[10pt, a4paper]{article}
\usepackage{lrec2022} % this is the new LREC2022 Style
\usepackage{multibib}
\newcites{languageresource}{Language Resources}
\usepackage{graphicx}
\usepackage{tabularx}
\usepackage{soul}
\usepackage{titlesec}
\titleformat{\section}{\normalfont\large\bfseries\center}{\thesection.}{1em}{}
\titleformat{\subsection}{\normalfont\SmallTitleFont\bfseries\raggedright}{\thesubsection.}{1em}{}
\titleformat{\subsubsection}{\normalfont\normalsize\bfseries\raggedright}{\thesubsubsection.}{1em}{}
\renewcommand\thesection{\arabic{section}}
\renewcommand\thesubsection{\thesection.\arabic{subsection}}
\renewcommand\thesubsubsection{\thesubsection.\arabic{subsubsection}}
\usepackage{epstopdf}
\usepackage[utf8]{inputenc}

\usepackage{hyperref}
\usepackage{xstring}

\usepackage{color}

\usepackage{booktabs}
\usepackage{amsmath,amsfonts}

\title{Data Augmentation with Paraphrase Generation and Entity Extraction for Multimodal Dialogue System}

\name{Eda Okur, Saurav Sahay, Lama Nachman} 

\address{Intel Labs \\
         USA \\
         \{eda.okur, saurav.sahay, lama.nachman\}@intel.com\\}

\abstract{
Contextually aware intelligent agents are often required to understand the users and their surroundings in real-time. Our goal is to build Artificial Intelligence (AI) systems that can assist children in their learning process. Within such complex frameworks, Spoken Dialogue Systems (SDS) are crucial building blocks to handle efficient task-oriented communication with children in game-based learning settings. We are working towards a multimodal dialogue system for younger kids learning basic math concepts. Our focus is on improving the Natural Language Understanding (NLU) module of the task-oriented SDS pipeline with limited datasets. This work explores the potential benefits of data augmentation with paraphrase generation for the NLU models trained on small task-specific datasets. We also investigate the effects of extracting entities for conceivably further data expansion. We have shown that paraphrasing with model-in-the-loop (MITL) strategies using small seed data is a promising approach yielding improved performance results for the Intent Recognition task.
 \\ \newline  }

\begin{document}

\maketitleabstract

\section{Introduction}

Building Artificial Intelligence (AI) systems that can assist students in the math learning process has been a challenging yet exciting area of research. Play-based learning systems can offer significant advantages in teaching fundamental mathematical concepts interactively, especially for younger kids~\cite{skene2021can}. Such intelligent systems are often required to handle multimodal understanding of the children and their surroundings in real-time. Within these complex architectures, Spoken Dialogue Systems (SDS) are crucial building blocks for carrying out efficient task-oriented communication with kids in game-based learning settings. This work presents our multimodal dialogue system for children learning basic math concepts. This dialogue system technology needs to be constructed and evaluated carefully to handle goal-oriented interactions between the kids and a virtual conversational agent. 

This study primarily focuses on creating and improving the Natural Language Understanding (NLU) module of a task-oriented SDS pipeline, especially with limited dataset resources. Building the NLU module of such goal-oriented SDS often involves the definition of intents (and entities if necessary), creation of domain-specific and task-relevant datasets, annotation of the data with intents and entities, iterative training and evaluation of NLU models, and repeating this process for each new or updated use-cases. This work explores the potential benefits of data augmentation with paraphrase generation for the NLU models trained on small-size task-specific datasets. The main NLU task we concentrate on improving is the Intent Recognition (IR) from possible user utterances. The ultimate goal of IR is to predict the user's intent (i.e., what the user wants to accomplish within a task-oriented SDS) given an input utterance. In addition to paraphrasing the possible user utterances for increased intent samples, we investigate the effects of extracting entities from these utterances for potentially further data expansion. 

Our experiments show that paraphrasing with model-in-the-loop (MITL) strategies is a promising approach to boost performance results for the IR task on our small-scale task-specific datasets. To be more precise, we first increase the F1-score of our baseline NLU model by 5\% (i.e., 90.6 to 95.6) for the intents by adopting a multi-task architecture. Then, we improve this further by 4\% (i.e., 95.6 to 99.4) with MITL data augmentation. With a Transformer-based multi-task architecture for joint Intent and Entity Recognition, we investigate employing and auto-annotating the entities to improve the NLU performance. Our next goal is to obtain more variations in paraphrased samples with entity expansion to create semantically richer datasets.

%This project is about building the Natural Language Understanding (NLU) modules of task-oriented Spoken Dialogue Systems (SDS). This process involves a definition of intents and entities, collection of task-relevant data, annotating the data with intents and entities, and then repeating the same process over and over again for adding any functionality/enhancement to the SDS. In this work, we showcase an Intent Bulk Labeling system where SDS developers and conversational designers can interactively label and augment training data from unlabeled utterance corpora using advanced clustering and visual labeling methods.

%\subsection{Related Work}
\section{Related Work}

%\cite{Authors-2018,Authors-2019,Authors-2021}
%\cite{KidSpace-ICMI-2018,KidSpace-SemDial-2019,KS2.0-NAACL-2021,KidSpace-ETRD-2022}

Investigating intelligent systems to assist children in their learning process has been an attractive area of research~\cite{MMHCI}. Employing Natural Language Processing (NLP) for building educational applications has also gained popularity in the past decade~\cite{meurers2012natural,lende2016question,taghipour2016neural,cahill-etal-2020-context}. Game-based learning environments can enhance significant benefits in teaching basic math concepts interactively, particularly for young learners~\cite{skene2021can}. Since we aim to build conversational agents for early childhood education with scarce datasets, we summarize the existing SDS/NLU approaches and data augmentation methods.% to boost the NLU performance with limited task-specific data.

%\subsubsection{SDS}
%\subsubsection{Conversational AI Systems}
\subsection{Conversational AI Systems}

Dialogue systems are often categorized as either task-oriented or open-ended, where the task-oriented dialogue systems are designed to fulfill specific tasks and handle goal-oriented conversations, whereas the open-ended systems or chat-bots allow more generic conversations such as chit-chat~\cite{Jurafsky:2000:SLP:555733}. With the advancements of deep learning-based language technologies and increased availability of large datasets in the research community, the end-to-end trained dialogue systems are shown to produce promising results for both goal-oriented~\cite{bordes2016learning} and open-ended~\cite{dodge2015evaluating} applications. Dialogue Managers (DM) of goal-oriented systems are often sequential decision-making models where optimal policies are learned via reinforcement learning from a high number of user interactions~\cite{shah2016interactive,dhingra2016towards,liu2017e2e,su-etal-2017-sample,cuayahuitl2017simpleds}. However, building such systems with limited user interactions is remarkably challenging. Thus, supervised learning approaches with modular SDS pipelines are still widely preferred when initial training data is limited, basically to bootstrap the goal-oriented conversational agents for further data collection~\cite{KidSpace-SemDial-2019}. Statistical and neural network-based dialogue system toolkits and frameworks~\cite{bocklisch2017rasa,ultes-etal-2017-pydial,burtsev-etal-2018-deeppavlov} are also heavily used in the academic and industrial research communities for implicit dialogue context management. A recent study named Conversation Learner~\cite{shukla-etal-2020-conversation} describes an interactive DM via machine teaching with human-in-the-loop annotations. Although the majority of task-oriented dialogue systems require defining intents and entities, a recent work called SMCalFlow~\cite{10.1162/tacl_a_00333} argues for a richer representation of dialogue state as a dataflow graph.

%\subsubsection{NLU}
%\subsubsection{Natural Language Understanding}
\subsection{Natural Language Understanding}

The NLU module within SDS processes user utterances as input and typically predicts the user intents and entities of interest. Long Short-Term Memory (LSTM) networks~\cite{lstm-1997} and Bidirectional LSTMs (BiLSTM)~\cite{bi-lstm-1997} have been widely used for sequence learning tasks such as Intent Classification~\cite{hakkani-2016} and Slot Filling~\cite{mesnil-2015}. Jointly training Intent Recognition and Entity Extraction models have been explored recently~\cite{zhang-2016,DBLP:journals/corr/LiuL16d,goo-etal-2018-slot,varghese2020bidirectional}. Various hierarchical multi-task architectures are also proposed for these joint NLU tasks~\cite{h-zhou-2016,wen-2018,AMIE-CICLing-2019,vanzo-etal-2019-hierarchical}, even some in multimodal context~\cite{gu2017speech,okur-etal-2020-audio}. \newcite{DBLP:conf/nips/VaswaniSPUJGKP17} proposed the Transformer, a novel network architecture based entirely on attention mechanisms~\cite{DBLP:journals/corr/BahdanauCB14}. Transformer-based models usually achieve better results than RNN-based models for the NLU tasks~\cite{KidSpace-SemDial-2019}. Bidirectional Encoder Representations from Transformers (BERT)~\cite{devlin2018bert} is one of the main breakthroughs in pre-trained language representations, showing strong performance in numerous NLP tasks, including the NLU. More recently,~\newcite{DIET-2020} introduced the Dual Intent and Entity Transformer (DIET), a lightweight multi-task architecture for joint Intent Classification and Entity Recognition. The authors showed that DIET outperforms fine-tuning BERT for predicting intents and entities on a complex multi-domain NLU-Benchmark dataset~\cite{Liu2021} and is much faster to train.

%\subsubsection{Data Augmentation}
\subsection{Data Augmentation}

Earlier studies have shown that data augmentation can improve the classification performance for several NLP tasks~\cite{barzilay-mckeown-2001-extracting,dolan-brockett-2005-automatically,DBLP:journals/corr/abs-1708-00391,hu-etal-2019-large}. Back-translation~\cite{sennrich-etal-2016-improving} is a popular method to construct paraphrase corpora by translating the samples to another language and then back to the original language.~\newcite{wieting-etal-2017-learning} employed the Neural Machine Translation (NMT)~\cite{DBLP:journals/corr/SutskeverVL14} approach for translating the non-English part of the parallel corpus to obtain English-to-English paraphrase pairs. Later, the authors massively scaled this approach to generate huge paraphrase corpora called ParaNMT-50M~\cite{wieting-gimpel-2018-paranmt}. In order to learn how to generate meaningful paraphrases, some of the previous work have utilized the autoencoders~\cite{10.5555/2986459.2986549,bowman-etal-2016-generating}, encoder-decoder models such as BART~\cite{DBLP:journals/corr/abs-1910-13461}, and NMT~\cite{sokolov2020neural}. Recent studies explore data augmentation via Natural Language Generation (NLG) for few-shot intents~\cite{xia-etal-2020-composed} and paraphrase generation for intents and slots in task-oriented dialogue systems~\cite{jolly-etal-2020-data}. Another relevant recent work~\cite{panda-etal-2021-multilingual} is an extension of a transformer-based model by~\newcite{jolly-etal-2020-data} that works for multilingual paraphrase generation for intents and slots, even in the zero-shot settings. Several other recent works have also been exploring data augmentation with fine-tuning large language models and few-shot learning for intent classification and slot-filling tasks~\cite{kumar-etal-2019-closer,kumar-etal-2020-data,DBLP:journals/corr/abs-2102-01335}.

%Previous work has shown that data augmentation can boost performance on text classification tasks~\cite{barzilay-mckeown-2001-extracting,dolan-brockett-2005-automatically,DBLP:journals/corr/abs-1708-00391,hu-etal-2019-large}.~\newcite{wieting-etal-2017-learning} used Neural Machine Translation (NMT)~\cite{DBLP:journals/corr/SutskeverVL14} to translate the non-English side of the parallel text to get English-English paraphrase pairs. This method has been scaled to generate large paraphrase corpora~\cite{wieting-gimpel-2018-paranmt}. Prior work in learning paraphrases has used autoencoders~\cite{10.5555/2986459.2986549}, encoder-decoder architectures as in BART~\cite{DBLP:journals/corr/abs-1910-13461}, and other learning frameworks such as NMT~\cite{sokolov2020neural}. Data augmentation using paraphrasing is a simple yet effective strategy that we explored in this work to improve the clustering. 

\section{Application Domain}
\label{app}
%\subsection{Application Domain}
%\subsection{Use-Cases}

%\begin{figure}[!h]
%\begin{figure*}[!ht]
\begin{figure*}[t]
\begin{center}
\includegraphics[scale=0.26]{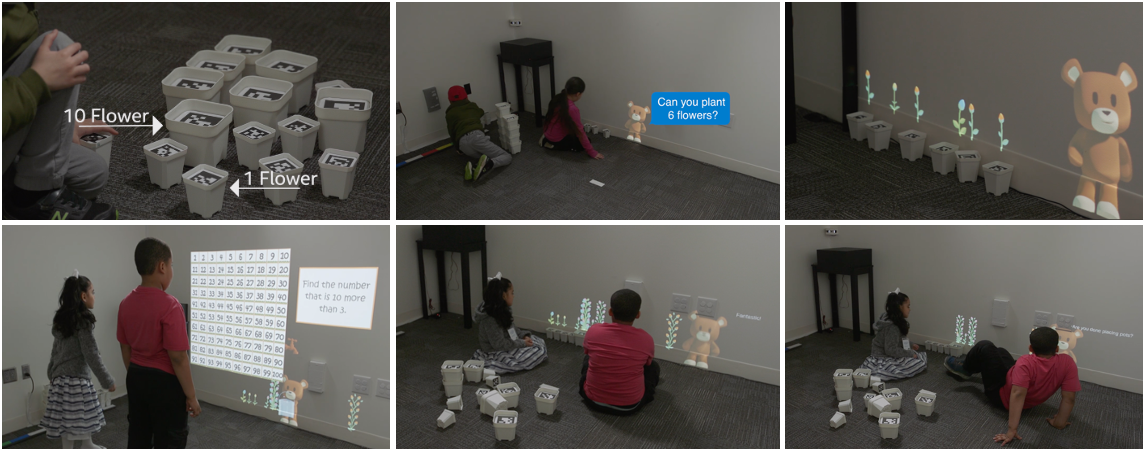} 
\caption{Learning basic math via play-based interactions.}
\label{fig1-use}
\end{center}
%\end{figure}
\end{figure*}

The motivation behind this study is to build conversational agents as part of the Kid Space project for early childhood education. Kid Space aims to create smart spaces for children with traditional gaming motivations such as level achievements and virtually collecting objects. The space allows multiple children to interact, which can encourage social development. The agent should accurately comprehend inputs from children and provide feedback. The system needs to be physically grounded to allow children to bring meaningful objects into the play experience, such as physical toys and manipulatives as learning materials. Thus, Kid Space AI system would combine various sensing technologies that should interact with children, track each child, and monitor their progress~\cite{KidSpace-ICMI-2018,KidSpace-ETRD-2022}.

The use-cases contain specific flows of interactive games facilitating elementary math learning designed for children 5-to-8 years old. The Flowerpot game builds on the math concepts of tens and ones, with the larger flowerpots representing `tens' and smaller pots `ones'. The virtual character provides a number of flowers the children should plant, and when the children have placed the correct number of pots against the wall, digital flowers appear. In the NumberGrid game, children are shown math clues, and when the correct number is touched on the grid, water is virtually poured to water the flowers. These experiences require the space to be physically grounded. As an example, at certain point in the game, the virtual character needs help to get on the table, where children position physical boxes that the character can jump upon, showing an understanding of the physical space.

Figure~\ref{fig1-use} demonstrates our intelligent agent, Oscar the teddy bear, helping the kids with learning tens and ones concepts along with practicing simple counting, addition, and subtraction operations.

The technology behind Kid Space includes distinct recognition capabilities. In the Flowerpot game, the 2D computer vision algorithm based on AprilTags is used to associate the physical flowerpots with the virtual flowers using a standard RGB camera. Our dialogue system takes multimodal information to incorporate user identity, actions, gestures, and the audio context. During the Flowerpot experience, the virtual character asks the children if they are done placing pots, to which they respond `yes'. Our dialogue system needs to use the visual input for the agent to respond appropriately to the correct number of pots being detected. The visual, audio, and LiDAR-based gesture recognition enables physically situated interactions. User identification utilizing the body and face would allow to accurately recognize children across different cameras and attribute their actions accordingly. 3D scene understanding algorithms are being used for the boxes experience to provide detection and 6D pose estimation of multiple concurrent boxes.

\section{Multimodal SDS}
%\section{Improving NLU for Multimodal SDS}

In this section, we describe the overall architecture of our multimodal dialogue system that we build for the Kid Space project. The aim is to enable children to interact with the agent while performing various activities, including learning math concepts with physical manipulatives or objects. For that, the dialog system must accurately comprehend multimodal inputs from children and respond appropriately.

Note that the current use-cases are designed for two children playing and learning with the agent collaboratively, while an adult user (i.e., Facilitator) is also present in the room to interact with the agent for game progress and helping the kids whenever needed. Therefore, we need to build a robust multimodal and multiparty conversational system that incorporates number of learning modules for interacting with multiple users (i.e., two children and one adult). This goal-oriented dialogue system should provide game instructions, guide the kids, understand both the kids' and Facilitator's utterances and actions to respond appropriately.

\subsection{Architecture}

%\begin{figure}[!h]
%\begin{figure*}[!ht]
\begin{figure*}[t]
\begin{center}
\includegraphics[scale=0.26]{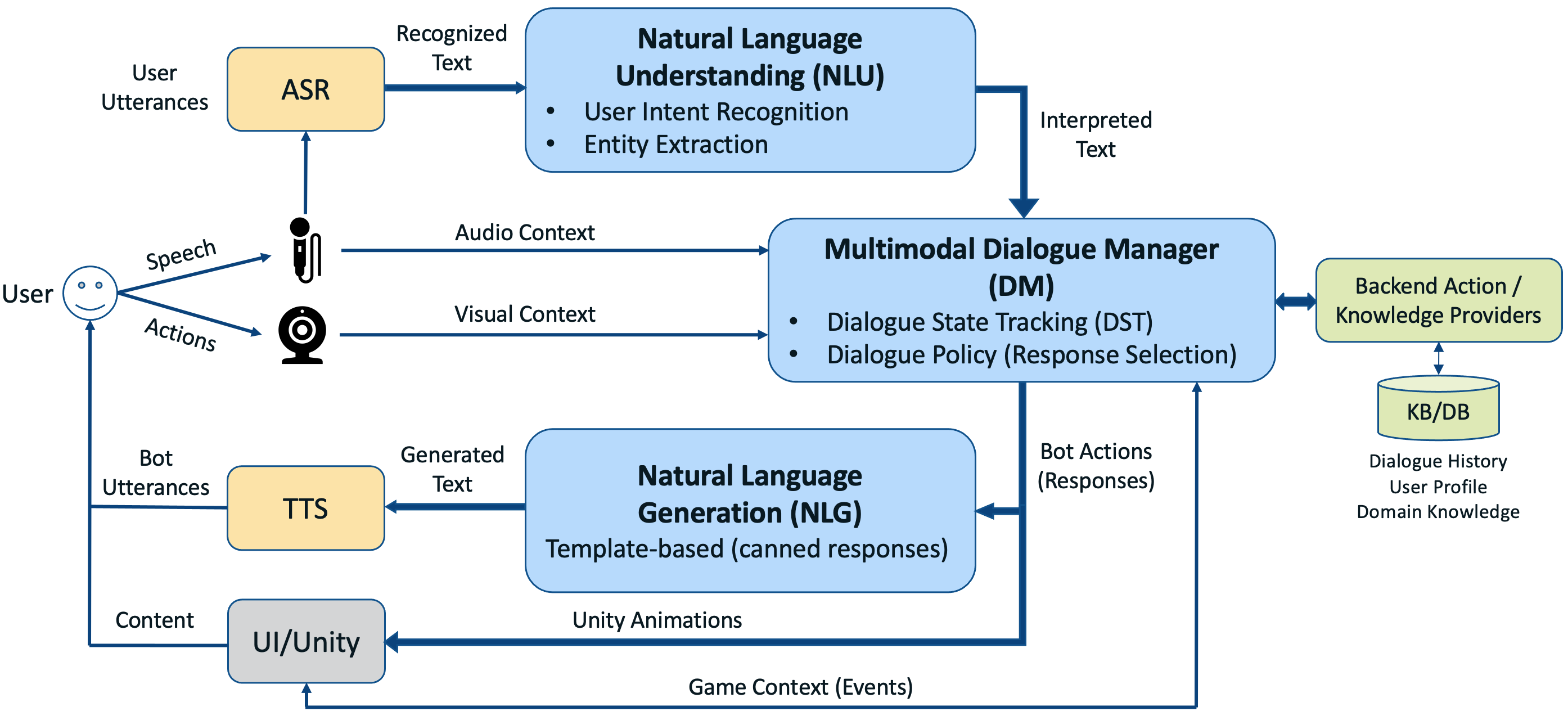} 
\caption{Multimodal SDS pipeline.}
\label{fig2-arch}
\end{center}
%\end{figure}
\end{figure*}

The SDS pipeline starts from recognizing user speech via Automatic Speech Recognition (ASR) module and feed the recognized text into our NLU component. We develop NLU models performing Intent Recognition and Entity Extraction to interpret user utterances. Then we pass these user intents and entities together with multimodal inputs, such as user actions and objects, into the DM component. The multimodal dialogue manager handles verbal and non-verbal communication inputs from the NLU (e.g., intents and entities) and separate external nodes processing visual and audio information (e.g., faces, poses, gestures, objects, events and actions). Note that we pass these multimodal inputs directly to the DM (bypassing the NLU) in the form of relevant multimodal intents for goal-oriented interactions. The Dialogue State Tracking (DST) model tracks what has happened (i.e., the dialogue state) within the DM. Then, the output of DST is used by the Dialogue Policy to decide which action the system should take next. Our DM models predict the appropriate agent actions and responses based on all the available contextual information (i.e., language-audio-visual inputs, game events, and dialogue history/context from previous turns). That means the DM generates sequential actions for bot utterances and non-verbal events. When the verbal response types are predicted, based on the output classes of DM, the NLG module retrieves actual bot responses, which are template-based in our use-cases. We create a variety of responses by preparing multiple templates (i.e., 3-to-6 variations) for each response type. Among these, the final response template randomly assigned at run-time. Generating grammatically correct and semantically coherent responses is challenging with such scarce datasets. Hence, this approach is more reliable than training NLG models for our application. Finally, the generated text responses are sent to the Text-to-Speech (TTS) module to output agent utterances. Non-verbal agent actions such as animations and game events are sent to the Unity application, which serves as the end User Interface (UI) displaying the agent and learning game content. Figure~\ref{fig2-arch} illustrates the schematic representation of our modular SDS pipeline. 

%Our application consists of interactions where children get introduced to the agent and they can play a simplified version of `Simon Says' game with the agent. The dialog manager ingests verbal and non-verbal communication via the Natural Language Understanding (NLU) component (entities and intents) and other engines that process vision and audio information (faces, pose, gesture, events and actions) and generates sequential actions for utterances and non-verbal events. We describe the NLU, Dialog Manager (DM) and {Dialog Adaptation} modules in this work as shown in Figure~\ref{fig:app}. We build our NLU and DM based models on top of the Rasa framework \cite{bocklisch2017rasa}. We enrich the NLU Intent Recognition model in Rasa by adding additional features to the model.

%Our application handles multimodal streams of high frequency non-verbal input such as person recognition via face and speaker identification, gestures, pose and audio events. We pass on the information via separate modules to the dialog manager (bypassing the NLU) as relevant intents for goal-oriented interaction. 

\section{Model Development}

This section describes the models we develop for the NLU (i.e., Intent classification and Entity Recognition) and multimodal DM modules of our dialogue system pipeline. For data augmentation explorations, we also discuss the development of our paraphrase generation models here.

\subsection{NLU and DM}
\label{nlu-models}

Our NLU and DM models are built on top of the Rasa open-source framework~\cite{bocklisch2017rasa}. The former baseline Intent Classifier in Rasa was based on supervised embeddings provided within the Rasa NLU, which is an embedding-based text classifier that embeds user utterances and intents into the same vector space. This former baseline architecture is inspired by the StarSpace work~\cite{wu2017starspace}, where the embeddings are trained by maximizing the similarity between intents and utterances. In previous work~\cite{KidSpace-SemDial-2019}, we enriched this former baseline NLU/Intent Recognition architecture available in Rasa by incorporating additional features and adapting alternative network architectures. To be more precise, we adapted the Transformer network~\cite{DBLP:conf/nips/VaswaniSPUJGKP17} and incorporated pre-trained BERT embeddings~\cite{devlin2018bert} to improve the Intent Recognition performance, as shown in~\newcite{KidSpace-SemDial-2019}. In this current study, our new baseline NLU model is the best-performing approach from our previous work~\cite{KidSpace-SemDial-2019}, which we would call TF+BERT in our experiments.

In this work, we explore potential improvements in Intent Classification performance by adapting the recent DIET architecture~\cite{DIET-2020}. DIET is a transformer-based multi-task architecture for joint Intent Recognition and Entity Extraction. DIET can incorporate pre-trained word and sentence embeddings from language models as dense features, with the flexibility to combine these with token level one-hot encodings and multi-hot encodings of character n-grams as sparse features. Note that one can use any pre-trained embeddings as dense features in DIET, such as GloVe~\cite{pennington2014glove}, BERT~\cite{devlin2018bert}, and ConveRT~\cite{ConveRT-2020}. Conversational Representations from Transformers (ConveRT) is another recent and promising architecture to obtain pre-trained representations that are well-suited for Conversational AI applications, especially for the Intent Classification task. Both DIET and ConveRT are lightweight architectures with faster training capabilities than their counterparts. For all the above reasons, we adapted the DIET architecture and incorporated pre-trained ConveRT embeddings to improve our Intent Classification performance (and later explore the Entity Recognition capabilities). We would call this approach DIET+ConveRT in our experiments\footnote{Please refer to the~\newcite{DIET-2020} for hyperparameters, hardware specifications, and computational cost details.}.

Although the DM model development is beyond the scope of this work, we gradually migrated from the baseline Recurrent Embedding Domain Policy (REDP)~\cite{vlasov2018few} model, which is again inspired by the StarSpace algorithm~\cite{wu2017starspace} and used in our previous work~\cite{KidSpace-SemDial-2019}. We adopted a more recent and suitable Transformer Embedding Dialogue (TED) policy~\cite{TED-2019} architecture, where a transformer's self-attention mechanism operates over the sequence of dialogue turns.

%SPACY? CRF?

\subsection{Paraphrase Generation}
\label{para}
%\subsection{Data Augmentation}

Data augmentation via paraphrase generation is an effective strategy that we explored in this study to improve the Intent Classification performance, particularly when we have limited original data to train our NLU models. With that motivation, we developed a data augmentation module via training a sequence-to-sequence paraphrasing model to generate paraphrased samples from the original seed utterances to augment the NLU training data. We propose several paraphrasing-based approaches and augmentation strategies to over-sample the intent classes and investigate their effects on the NLU performance. We examine the data augmentation via paraphrasing with a few simple heuristics (i.e., paraphrasing only for the low-sample intents or minority classes or excluding the intent types with samples having shorter utterance lengths). We also investigate model-in-the-loop data augmentation techniques (i.e., augmenting only the paraphrased utterances with successful predictions and checking the confidence level thresholds using the initial NLU models trained on original samples).

For effective paraphrase generation from the seed samples, we adapted the BART sequence-to-sequence model~\cite{DBLP:journals/corr/abs-1910-13461} that we fine-tuned on the back-translated English sentences from the combination of following three datasets: the Microsoft Research Paraphrase (MSRP) corpus~\cite{dolan-brockett-2005-automatically}, ParaNMT corpora~\cite{wieting-gimpel-2018-paranmt,wieting-etal-2017-learning}, and the PAWS dataset~\cite{paws2019naacl,pawsx2019emnlp}. The MSRP is a corpus containing 5800 pairs of sentences extracted from web news sources, along with human annotations indicating whether each pair captures a semantic equivalence/paraphrase relationship. The ParaNMT-50M is a dataset of more than 50 million English-English sentential paraphrase pairs back-translated from the Czeng1.6 corpus~\cite{bojar2016czeng}. The PAWS is a corpus containing 108,463 human-labeled and well-formed paraphrase pairs. Note that we also trained paraphrasing models using GPT-2~\cite{radford2019language} fine-tuning to augment the training set. Finally, we decided to stick with the sequence-to-sequence paraphrasing model with BART fine-tuning as it performed slightly better than GPT-2 fine-tuning version, which is expected as the BART model can be seen as a generalized BERT~\cite{devlin2018bert}  encoder and GPT~\cite{radford2018improving} decoder.

%Data augmentation using paraphrasing is a simple yet effective strategy that we explored in this work to improve the clustering. 

%For handling data imbalance, we propose a paraphrasing-based method to over-sample the minority classes. The method is described as follows:

%For paraphrase generation, we utilized the BART Sequence-to-Sequence model~\cite{DBLP:journals/corr/abs-1910-13461} fine-tuned on the combination of following three datasets: the MSRP corpus~\cite{dolan-brockett-2005-automatically}~\citelanguageresource{MSRP}, ParaNMT corpora~\cite{wieting-gimpel-2018-paranmt,wieting-etal-2017-learning}~\citelanguageresource{ParaNMT}, and the PAWS dataset~\cite{paws2019naacl,pawsx2019emnlp}~\citelanguageresource{PAWS}.

\begin{table}[!t]
%\begin{table*}[!h]
  \centering
  \small
  %\resizebox{\columnwidth}{!}{
  \begin{tabular}{lcc}
    \toprule
    \textbf{Statistics/Dataset} & \textbf{Planting} & \textbf{Watering} \\
    \midrule
    \# distinct intents & 14 & 13 \\
    total \# samples (utterances) & 1927 & 2115 \\
    min \# samples per intent & 22 & 25 \\
    max \# samples per intent & 555 & 601 \\
    avg \# samples per intent & 137.6 & 162.7 \\
    \# unique words (vocab) & 1314 & 1267 \\
    total \# words & 10141 & 10469 \\
    min \# words per sample & 1 & 1 \\
    max \# words per sample & 74 & 65 \\
    avg \# words per sample & 5.26 & 4.95 \\
    \bottomrule
  \end{tabular}
  %}
  \caption{KidSpace NLU Dataset Statistics}
  \label{data-st}
\end{table}
%\end{table*}

\section{Experimental Results}

\subsection{Dataset}

We conduct our experiments on the KidSpace NLU datasets having utterances from multimodal math learning experiences (i.e., Planting and Watering activities) designed for 5-to-8 years-old kids~\cite{KS2.0-NAACL-2021,KidSpace-ETRD-2022}. These are the initial proof-of-concept (POC) datasets to bootstrap the agents to be deployed. These POC datasets are manually created based on User Experience (UX) design studies for training the SDS models and validated with UX sessions in the lab with multiple kids going through play-based learning activities. The NLU datasets have a limited number of utterances, which are manually annotated for intent types that we defined for each use-case or learning game/activity (see section~\ref{app}). For the Flowerpot game, we have the Planting Flowers dataset with 1927 utterances, and for the NumberGrid activity, we have a separate Watering Flowers dataset with 2115 utterances. Some of our intents are highly generic across usages and activities (e.g., \textit{affirm}, \textit{deny}, \textit{next\_step}, \textit{out\_of\_scope}, \textit{goodbye}), whereas the rest are highly domain-dependent and task-specific (e.g., \textit{intro\_meadow}, \textit{answer\_flowers}, \textit{answer\_water}, \textit{ask\_number}, \textit{counting}). Note that our dialogue system needs to process and interpret utterances received either from the kids or the adult (i.e., the Facilitator) present in the room. Therefore, almost half of our intent types are defined based on what the game flow expects from the Facilitator (e.g., to progress the games or guide the children). Table~\ref{data-st} shows the statistics of our NLU datasets.

%To conduct our experiments, we use the KidSpace dataset that includes utterances from a Multimodal Learning Application for 5-to-8 years-old children~\cite{KS2.0-NAACL-2021,KidSpace-SemDial-2019,KidSpace-ICMI-2018}. Table 1 in~\newcite{KS2.0-NAACL-2021} shows the statistics of the datasets.
%Table~\ref{data-st} shows the statistics of the datasets.

\subsection{NLU Results}

To evaluate the Intent Recognition performances, the baseline NLU model that we call TF+BERT is compared with the DIET+ConveRT model that we adapted recently (see section~\ref{nlu-models}). We conduct these evaluations on both the Planting and Watering datasets. Table~\ref{nlu-results} summarizes the Intent Classification performance results in weighted average F1-scores. We perform 10-fold cross-validation (CV) over the dataset for each run, and we report the results based on the average of 3 runs.

\begin{table}[!t]
  \centering
  \small
  \begin{tabular}{lcc}
    \toprule
    \textbf{Model/Dataset} & \textbf{Planting} & \textbf{Watering} \\
    \midrule
    %TF+BERT (Baseline) & 90.6 & 92.4 \\
    TF+BERT (Baseline) & 90.55 & 92.41 \\
    %DIET+ConveRT & 95.6 & 97.8 \\
    DIET+ConveRT & \textbf{95.59} & \textbf{97.83} \\
    \midrule
    %Performance Gain & +5.0 & +5.4 \\
    Performance Gain & +5.04 & +5.42 \\
    \bottomrule
  \end{tabular}
  \caption{NLU/Intent Recognition F1-scores (\%): Previous (TF+BERT) and Updated (DIET+ConveRT) Model Results (3 runs of 10-fold CV)}
  \label{nlu-results}
\end{table}

As we can observe from Table~\ref{nlu-results}, adapting the lightweight DIET architecture~\cite{DIET-2020} with pre-trained ConveRT embeddings~\cite{ConveRT-2020} for our NLU models significantly improved the Intent Classification performance, which is consistent across different use-cases (i.e., Planting and Watering Flowers game datasets). With that observation, we have updated the NLU component in our multimodal SDS pipeline (see Figure~\ref{fig2-arch}) by replacing the TF+BERT model (i.e., previously best-performing Baseline + BERT + Transformer model in~\newcite{KidSpace-SemDial-2019}) with this promising DIET+ConveRT model.

\subsection{Data Augmentation}

The goal here is to explore the potential benefits of paraphrasing-based data augmentation to further improve the Intent Recognition models. Our final BART-based Seq2Seq model\footnote{\url{https://simpletransformers.ai/docs/seq2seq-model/}} (see section~\ref{para}) is utilized for paraphrase generation to conduct augmentation experiments. All paraphrasing experiments are conducted on the Planting Flowers dataset with a limited number of original seed utterances (i.e., less than 2K samples). We propose and investigate the following data augmentation methods with certain rule-based heuristics and model-in-the-loop (MITL) approaches:

\begin{itemize}
\item{\textbf{baseline (aug3/aug5/aug10)}: Augment the original NLU dataset with paraphrased samples. We configured the number of samples to be generated as 3/5/10 (i.e., for each original utterance, x3/x5/x10 paraphrased samples are generated).}
\item {\textbf{inc6low}: Augment data only for 6 low-sample intents (i.e., having less than 50 utterances). We simply used the original plus paraphrased samples for 6 intents with fewer number of utterances (i.e., \textit{intro\_meadow}, \textit{help\_affirm}, \textit{everyone\_understand}, \textit{oscar\_understand}, \textit{ask\_number}, \textit{next\_step}), whereas we only used the original samples for rest of the 8 intents with higher number of utterances (no need for more variations for those).}
\item {\textbf{exc5short}: Augment data except for 5 intents with seed samples having short utterance lengths. We only used the original samples for 5 intents with short utterances (i.e., \textit{affirm}, \textit{deny}, \textit{answer\_flowers}, \textit{answer\_valid}, \textit{answer\_invalid}), whereas we used the original plus paraphrased samples for rest of the 9 intents with longer utterances (as variations help for those).}
\item {\textbf{success}: Augment only the paraphrased samples that are classified correctly into the same intent class as their seed samples (successful predictions). For this MITL approach, we first trained the NLU model (DIET+ConveRT) on the original/seed dataset, then classified the paraphrased samples using this initial NLU model to obtain successful predictions. We assume the paraphrased samples should belong to the same class as seed samples, and the idea is to filter out noisy synthetic samples that belong to other classes.}
\item {\textbf{success\_conf90}: Augment only the paraphrased samples that are classified correctly into the same intent class as their seed samples (successful predictions) with a confidence score of 0.9 or higher. Another MITL approach using the same initial NLU model as in \textit{success}. The confidence check ensures better noise filtering, and the threshold of 0.9 is chosen empirically after checking the confidence histograms on paraphrased samples.}
\item {\textbf{all\_conf90}: Augment only the paraphrased samples that are classified into any intent type (regardless of their seed samples' intent class) with a confidence score of 0.9 or higher. Another MITL approach using the same initial NLU model as in \textit{success}. We removed the assumption that paraphrased samples should be of the same class as their seed samples and still augment them into the predicted class samples if confidence is high.}
\end{itemize}

\begin{figure}[b]
\begin{center}
\includegraphics[scale=0.25]{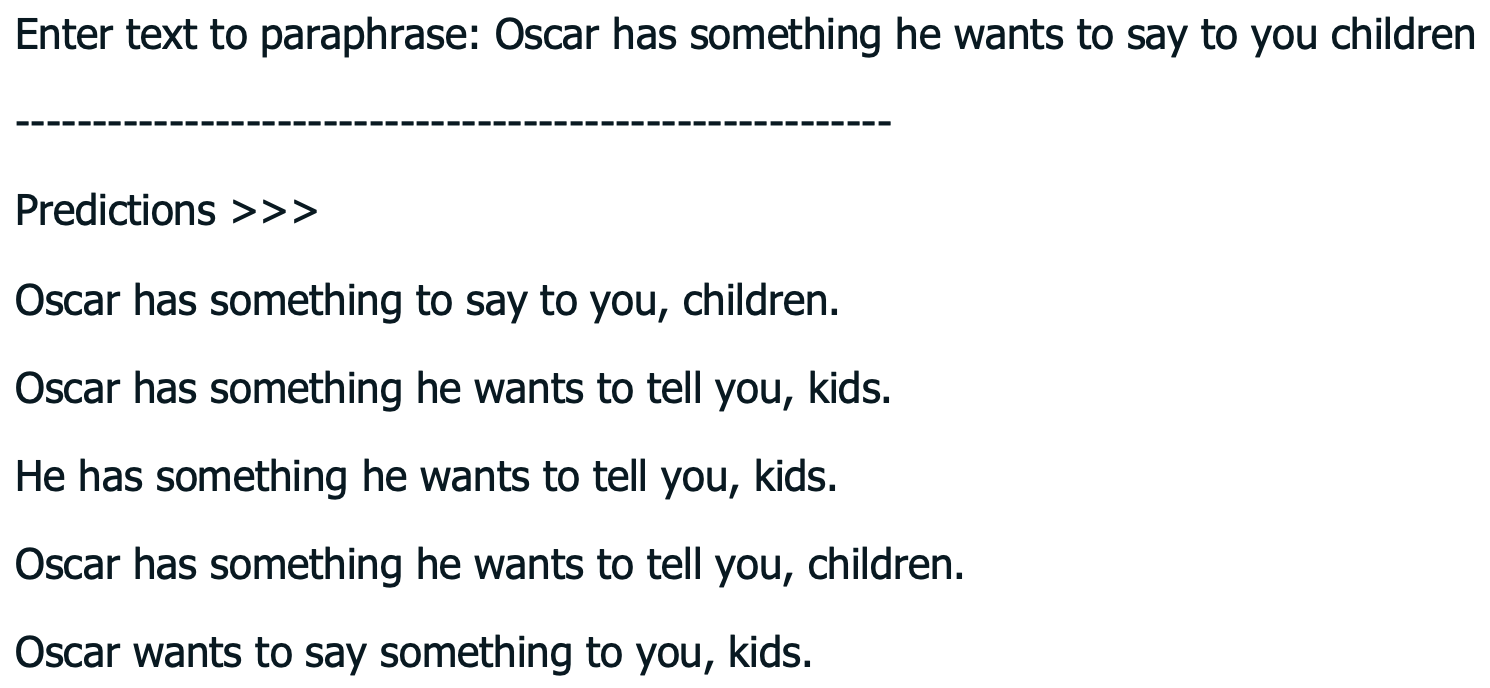}
\caption{Example seed and paraphrased utterances.}
\label{fig.ex}
\end{center}
\end{figure}

Figure~\ref{fig.ex} depicts an example seed utterance and its paraphrase generation outputs for x5. In all data augmentation experiments, we observed repetitions due to duplicate samples generated by the paraphraser, and we removed those duplicates from the final augmented datasets. Note that while augmenting the data, paraphrased samples are assumed to have the same intent labels as their seed samples they are generated from, except for the \textit{all\_conf90} case (for which we assign the labels predicted by the initial NLU model if the confidence is 0.9 or higher).

Table~\ref{para-results} summarizes the Intent Recognition performances in weighted-avg F1-scores after the data augmentation with paraphrased samples. We report the NLU results on an average of 3 runs and perform a 10-fold CV over original/augmented datasets for each run. In each such fold, the 10\% test partition has the original samples only, whereas the 90\% training partition is augmented with the paraphrased samples. This setup ensures we evaluate the models on the same original/seed samples only, while we can expand the training data with more variations.
As shown in Table~\ref{para-results}, the baseline approach of simply augmenting the data with paraphrased samples does not help but slightly hurt the NLU performances. That is due to the possible noises in synthetic data generated via paraphraser. However, with our proposed heuristics and MITL approaches, data augmentation helps improve our NLU results. When we augment for low-sample intents only (i.e., \textit{inc6low}), we start observing slight improvements and increasing the number of paraphrased samples help. We get event better jumps when we exclude short-sample intents (i.e., \textit{exc5short}). Augmenting only the paraphrased samples that are classified (with high confidence) into the same label as their seed samples (i.e., \textit{success\_conf90}) is the best performing approach, boosting the performance by nearly 4\% (compared to training on original samples). Since we properly filter out noisy synthetic data in this case, generating more paraphrases beyond x3 and x5 also helps (i.e., aug10 $>$ aug5 $>$ aug3 for \textit{success\_conf90}).

\begin{table}[!t]
  \centering
  \small
  \begin{tabular}{lcccc}
    \toprule
    \textbf{Method} & \textbf{original} & \textbf{aug3} & \textbf{aug5} & \textbf{aug10} \\
    \midrule
    %baseline & \textbf{95.6} & 95.2 & 94.7 & 94.8 \\
    baseline & \textbf{95.59} & 95.17 & 94.73 & 94.75 \\
    \midrule
    %inc6low & - & 95.8 & 96.1 & 96.4 \\
    inc6low & - & 95.84 & 96.06 & 96.41 \\
    %exc5short & - & 97.7 & 97.6 & 97.9 \\
    exc5short & - & 97.70 & 97.61 & 97.86 \\
    \midrule
    %success & - & 98.6 & 98.8 & 98.6 \\
    success & - & 98.58 & 98.82 & 98.65 \\
    %success\_conf90 & - & \textbf{99.2} & \textbf{99.4} & \textbf{99.4} \\
    success\_conf90 & - & \textbf{99.19} & \textbf{99.37} & \textbf{99.43} \\
    %all\_conf90 & - & 98.6 & 98.8 & 98.6 \\
    all\_conf90 & - & 98.61 & 98.75 & 98.58 \\
    \midrule
    Perf. Gain & - & +3.60 & +3.78 & +3.84 \\
    \bottomrule
  \end{tabular}
  \caption{NLU/Intent Recognition F1-scores (\%) with DIET+ConveRT Models Trained on Augmented Data via Paraphrasing}
  \label{para-results}
\end{table}

%\begin{figure}[!h]
%\begin{figure*}[!ht]
%\begin{figure*}[t]
\begin{figure}[h]
\begin{center}
\includegraphics[scale=0.17]{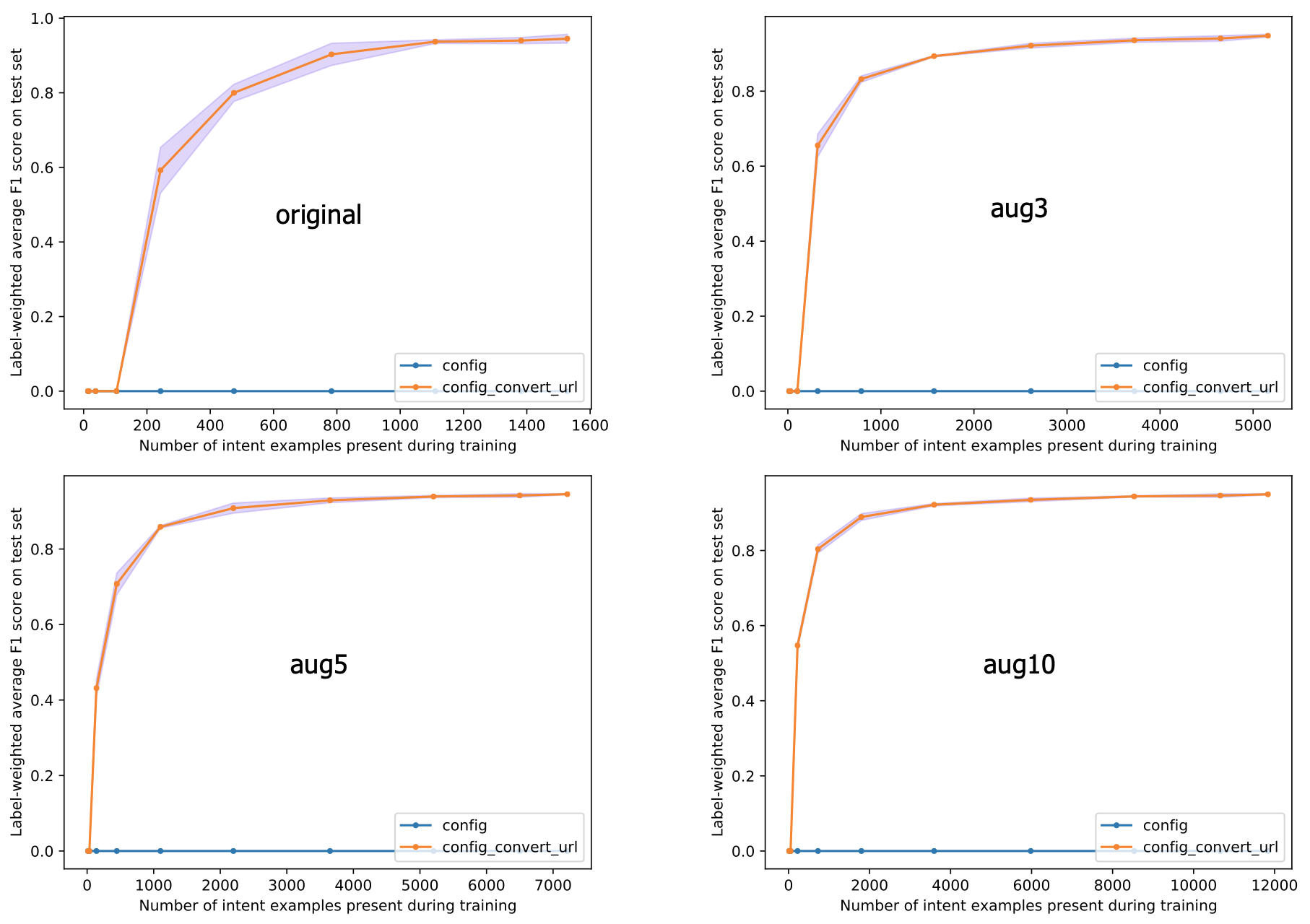} 
\caption{Paraphraser evaluation: data size vs. NLU performance for original and augmented datasets.}
\label{fig3}
\end{center}
\end{figure}
%\end{figure*}

%\begin{figure}[!b]
%\begin{figure*}[!ht]
%\begin{figure*}[h]
\begin{figure}[b]
\begin{center}
\includegraphics[scale=0.22]{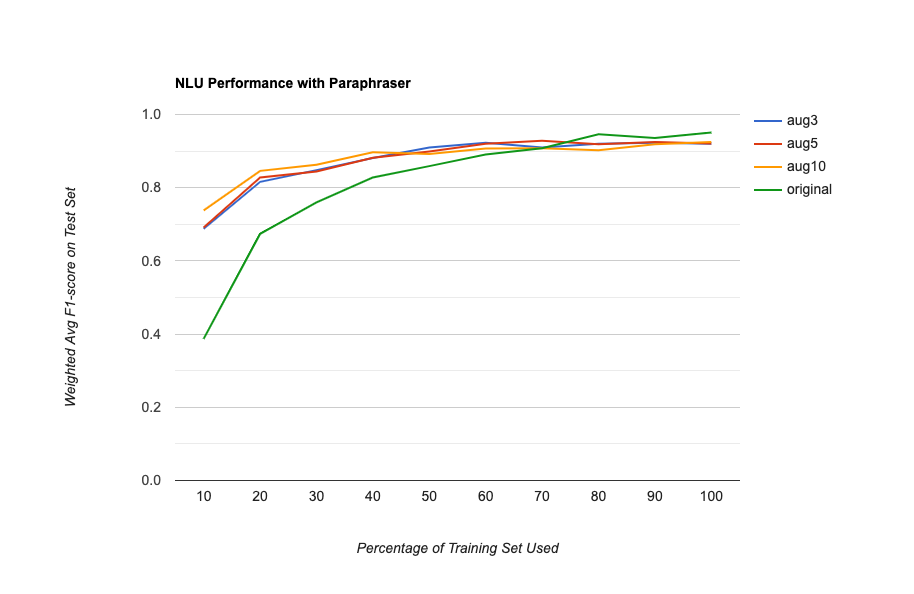}
\caption{Paraphraser evaluation: data percentage vs. NLU performance for original and augmented datasets.}
\label{fig4}
\end{center}
\end{figure}
%\end{figure*}

\subsubsection{Data Size versus NLU Performance}

We analyzed the dataset sizes vs. Intent Recognition F1-scores for original and augmented datasets with paraphrased samples. Instead of performing a 10-fold CV, we created a train/test split (i.e., 80/20\%) from each dataset (i.e., original, aug3, aug5, aug10). Then we trained NLU models with 10, 20, …, 100\% of the training sets, each evaluated on the same test set, and the F1-scores compared. This process repeated for 3 runs (i.e., 3 test sets). In Figure~\ref{fig3}, we show the plots for data size versus NLU performance with avg/std of F1-scores. Note that x-axis values (i.e., number of training samples) at each plot vary due to data augmentation. We observed that models trained on augmented datasets achieve a plateau of F1-scores faster with less original training data.

Next, we visualize the data size vs. performance with a superimposed chart for better comparison. We created a fixed train/test split (i.e., 80/20\%) from the original data. Then, we used the same test set across all comparisons (models can be trained on original/augmented data but tested only on original samples). For training sets, we created 10, 20, …, 100\% of the original training set, and we augmented those with the paraphrased samples (aug3, aug5, aug10). In Figure~\ref{fig4}, we plot the superimposed chart with common x-axis values (\% of the original training set used) for comparison.
We observed that we could reach a 0.8 F1-score with around 15\% of the original samples with paraphrasing (aug10), whereas we need at least 35\% of the data to achieve the same level of performance without paraphrasing (i.e., 2.3x reduction in required original data). Similarly, we can reach a 0.9 F1-score with around 40\% of the original set via paraphrasing (aug10), whereas we need at least 70\% of the data to achieve the same level of performance without paraphrasing (i.e., 1.75x reduction in required data). We believe this paraphrasing approach would help us achieve better results with limited initial intent samples whenever there is a new use-case (e.g., future learning activities in Kid Space).

\iffalse

We classify the paraphrased sample with a DIET~\cite{DIET-2020} based NLU model with ConveRT~\cite{ConveRT-2020} embeddings, which is trained on the original dataset. We only augment the paraphrased sample if the Intent Classifier predicts the same label as the seed instance. We name this the approach `success' in our experiments. Without this last step, we call this overall approach our `Paraphrasing' method.

We use the Paraphrasing model and the classifier as a data augmentation method to augment the labeled training data (refer to as `Aug' in our experiments).

Note that we augment the paraphrased sample if it belongs to the same minority class (`ParaMote') as we do not want to inject noise while solving the data imbalance problem. The opposite is also possible for other purposes such as generating semantically similar adversaries~\cite{ribeiro-etal-2018-semantically}.

Data Balancing

We apply Paraphrasing methods to balance the data. Paraphrasing almost always improves the performance while the additional classifier to check for class-label consistency does not help.  

Data Augmentation

We augmented the entire labeled data including the majority class using Paraphrasing (with class-label consistency) by 3x in our experiments. We aimed to understand if this could help get a better pre-trained model that could eventually improve the clustering outcome. We do not observe any performance gains with the augmentation process.

\fi

\subsection{Entity Extraction}

In addition to data augmentation, we aim to investigate the Entity Extraction and evaluate the potential improvements via entity expansion. The idea behind this is, if we can find a way to auto-extract the entities existing in the dataset, we can perform a lexical entity enrichment via ConceptNet~\cite{liu2004conceptnet,speer2017conceptnet} as an external Knowledge Graph (KG). Later on, we can also explore the use of lookup tables and synonyms in the dataset to create more variations via entity expansion on top of the original and paraphrased samples. With that motivation, we used a pre-trained SpaCy Entity Recognizer\footnote{\url{https://spacy.io/api/entityrecognizer}} that performs Named Entity Recognition (NER)~\cite{mccallum2003early,okur2016named} to automatically extract the entities in our original NLU dataset (Planting Flowers). We then re-formatted the dataset with auto-tagged entities detected within utterances. We performed 3 runs of 10-fold CV on the original dataset with SpaCy NER tagged entities. We evaluated the joint Intent and Entity Recognition performances using DIET+ConveRT.  

We observed that these auto-tagged entities generated via a pre-trained SpaCy NER model do not really help improve the NLU performances. We also realized that these generic named entity types are not very much relevant to our dataset. Hence, we had to define and extract more domain-specific entities, which requires a heavy task of word-level annotations. We completed these manual annotations for domain-specific entity types on the original dataset (Planting Flowers). 

Table~\ref{entity-results} summarizes our findings. Entity Recognition F1-score improved from 72.8\% to 97.1\% with these manual annotations, which is not surprising. Intent Classification F1-scores slightly drop when entities come into play, which aligns with the findings in~\newcite{DIET-2020}. Although we can extract domain-specific entities pretty accurately, these token-level manual annotations are costly, even for small-size datasets. Next, we investigate auto-annotating the domain-specific entities using ConceptNet relatedness. We provide up to 6 sample values for each domain-specific entity types that we previously defined, then construct a synonym dictionary by returning the corresponding entities in the KG if the relatedness of the value is larger than an empirical threshold of 0.7. After extracting the tokens and noun chunks via SpaCy part-of-speech (POS) tagger\footnote{\url{https://spacy.io/api/tagger}}, we automatically annotate the domain-specific entities. We do not expect this simplistic approach to work as accurately as manual annotations but want to see how much we can improve upon generic SpaCy NER tagged entities. Surprisingly, we achieved an Entity Recognition F1-score of 92.6\% with the auto-annotated entities, which we believe is a very good compromise.

We believe these domain-specific entities would help us achieve lexical entity enrichment via ConceptNet as a knowledge graph on top of the original/paraphrased samples. We also believe these results are quite encouraging in our quest to make dialog systems more robust and generalizable to new intents with limited data.  

%We notice that this representation not only helps with the labeling but also helps with correcting the labels and identify utterances that belong to multiple classes which cannot be easily segmented. For example, `children-valid-answer' and `children-invalid-grow' (invalid answers) contain semantically similar content depending on the game logic of the interaction. We perhaps need to group these together and use an alternative logic for implementing game semantics. 

\begin{table}[!t]
  \centering
  \small
  \resizebox{\columnwidth}{!}{
  \begin{tabular}{lcc}
    \toprule
     & \textbf{Intent} & \textbf{Entity} \\
     & \textbf{Classification} & \textbf{Recognition} \\
    \midrule
    %No entities & 95.6 & - \\
    No entities & \textbf{95.59} & - \\
    \midrule
    %SpaCy NER tags (auto) & 94.7 & 72.8 \\
    SpaCy NER tagged entities & 94.70 & 72.82 \\
    %Annotated entities (manual) & 94.9 & 97.1 \\
    Manually annotated entities & 94.91 & \textbf{97.12} \\
    \midrule
    %Performance Diff. & -0.68 & +24.3 \\
    Auto-annotated entities & 94.76 & 92.64 \\
    \bottomrule
  \end{tabular}
  }
  \caption{NLU/Joint Intent Classification and Entity Recognition F1-scores (\%): DIET+ConveRT Model Results (3 runs of 10-fold CV)}
  \label{entity-results}
\end{table}

\section{Conclusion and Future Work}

%TBD

Constructing robust dialogue systems are critical to achieving efficient task-oriented communication with children in game-based learning settings. This study presents our multimodal dialogue system engaging with younger kids while learning basic math concepts. We focus on improving the NLU module of the task-oriented SDS pipeline with limited datasets. This exploration employs data augmentation with paraphrasing to increase NLU performances. Paraphrasing with model-in-the-loop strategies looks promising for achieving higher F1-scores for Intent Classification using small task-dependent datasets. Finally, we investigate the Entity Extraction to potentially further improve the NLU component of our multimodal SDS.

In future work, we plan to extend the Plug and Play Language Model (PPLM)~\cite{dathathri2019plug} architecture applicable to the Decoder-only Unconditional Language Models (such as GPT-2) to Seq2Seq Encoder-Decoder based Conditional Language Models (CLM)~\cite{DBLP:journals/corr/abs-1909-05858} (such as BART), where the text to be generated is constrained by a cross-attention to the Encoder input. We can further control this CLM with controllable attributes that require no training/fine-tuning of the model. During inference, control attributes directly update the latent activations to steer the model to generate fluent and attribute-specific text. We can explore the PPLM approach to get more paraphrased samples using entity expansion via ConceptNet and adapt this approach to Seq2Seq Encoder-Decoder models.

\section{Acknowledgements}

%Place all acknowledgements (including those concerning research grants and funding) in a separate section at the end of the paper.

%Special thanks to Glen Anderson and the Anticipatory Computing Lab Kid Space team for conceptualization and the UX design for all the interactions. We thank Zhichao Hu (UCSC) who worked as a Summer Intern with us in 2017 and worked on the Dialog Adaptation module. We greatfully acknowledge and thank the Rasa Team and community developers for the framework and contributions that enabled us to further our research and build newer models for the application. 

%We want to show our gratitude to our colleagues from Intel Labs, especially, Cagri Tanriover for his tremendous efforts in coordinating and implementing the vehicle instrumentation to enhance multi-modal data collection setup (as he illustrated in Fig.~\ref{fig:car}), John Sherry and Richard Beckwith for their insights and expertise that guided the collection of this UX grounded and ecologically valid dataset (via scavenger hunt protocol and WoZ research design). The authors are also immensely grateful to the GlobalMe, Inc. members, especially Rick Lin and Sophie Salonga, for their extensive efforts in organizing and executing the data collection, transcription, and some annotation tasks for this research in collaboration with our team at Intel Labs.

We show our gratitude to our current and former colleagues from the Intel Labs Kid Space team, especially Ankur Agrawal, Glen Anderson, Sinem Aslan, Arturo Bringas Garcia, Rebecca Chierichetti, Hector Cordourier Maruri, Pete Denman, Lenitra Durham, Roddy Fuentes Alba, David Gonzalez Aguirre, Sai Prasad, Giuseppe Raffa, Sangita Sharma, and John Sherry, for the conceptualization and the design of use-cases to support this research. The authors are also immensely grateful to the Intel Labs KAIU team, specifically Nagib Hakim, Ezequiel Lanza, and Gadi Singer, for their feedback and support on entity annotation tasks in collaboration with our team. Finally, we thankfully acknowledge the Rasa team for the open-source framework and the community developers for their contributions that enabled us to improve our research and build proof-of-concept models for our use-cases.

\section{Bibliographical References}\label{reference}
%\label{main:ref}

\bibliographystyle{lrec2022-bib}
\bibliography{lrec2022-example}

%\section{Language Resource References}
%\label{lr:ref}
%\bibliographystylelanguageresource{lrec2022-bib}
%\bibliographylanguageresource{languageresource}

\end{document}